\documentclass[10pt,twocolumn,letterpaper]{article}
\usepackage[T1]{fontenc}
\usepackage[utf8]{inputenc}
\usepackage{newtxtext,newtxmath}
\usepackage[margin=0.72in,columnsep=0.25in]{geometry}
\usepackage{microtype}
\usepackage{graphicx}
\usepackage{amsmath}
\usepackage{booktabs,array,tabularx,multirow}
\usepackage{caption}
\usepackage{float}
\usepackage{algorithm}
\usepackage{algpseudocode}
\usepackage[numbers,sort&compress]{natbib}
\usepackage{enumitem}
\usepackage{placeins}
\usepackage{balance}
\usepackage{etoolbox}
\makeatletter
\patchcmd{\@BAlancecol}
  {\setbox\@leftcolumn=\vbox to \@tempdima}
  {\@tempdimb=\dp1\relax
   \ifdim\dp3>\@tempdimb\@tempdimb=\dp3\fi
   \advance\@tempdima by \@tempdimb
   \setbox\@leftcolumn=\vbox to \@tempdima}{}{}
\makeatother
\usepackage[hidelinks,pdfauthor={Pengcheng Pan and Xinfang Zhang},pdftitle={Q-CueGraph: Query-Conditioned Visual Evidence Graphs for Multimodal Reasoning}]{hyperref}
\newcommand{\qcg}{Q-CueGraph}
\newcommand{\vstar}{V*Bench}
\newcommand{\area}{\operatorname{area}}
\newcommand{\bbox}{\operatorname{bbox}}
\newcommand{\pad}{\operatorname{pad}}
\newcommand{\crop}{\operatorname{crop}}
\newcommand{\ANLS}{\operatorname{ANLS}}
\newcommand{\smalltable}{\small\setlength{\tabcolsep}{4.2pt}\renewcommand{\arraystretch}{1.10}}
\newcommand{\fullfigure}[3]{\begin{figure*}[t]\centering\includegraphics[width=\textwidth]{figures/#1}\caption{#2}\label{#3}\end{figure*}}
\begin{document}
\twocolumn[{
\begin{center}
{\LARGE\bfseries Q-CueGraph: Query-Conditioned Visual Evidence Graphs\\[2pt] for Multimodal Reasoning\par}
\vspace{8pt}
{\large Pengcheng Pan$^{1}$\qquad Xinfang Zhang$^{2}$\par}
\vspace{3pt}
{$^{1}$The University of Tokyo\qquad $^{2}$Tohoku University\par}
\vspace{2pt}
{\small\texttt{pan@isi.imi.i.u-tokyo.ac.jp}\quad\texttt{xinfang.zhang.b5@tohoku.ac.jp}\par}
\end{center}
\vspace{2pt}
\begin{minipage}{\textwidth}
\small
\textbf{Abstract.}\quad
Multimodal large language models (MLLMs) can miss fine details in a full image that they recognize in a closer view. Recovering this evidence requires deciding where to look and how much surrounding context to retain. We present \qcg, a query-conditioned evidence acquisition method for frozen MLLMs. For text-rich images, it builds a reusable graph of OCR lines and layout relations. Each question activates anchors, expands them into contextual regions, and selects candidates for a single observation window. Query-conditioned object detections support natural-image search through the same region-selection and composition interface. A lightweight candidate scorer further learns which observations support correct answers from frozen-reader feedback and training answers, without evidence-box supervision. Across six benchmarks, we examine the roles of query conditioning, evidence composition, and learned answerability. With Qwen2.5-VL-7B, \qcg\ raises \vstar\ accuracy from 0.696 to 0.832 using 19.1\% of source-image area, and retains 92\% of full-image ANLS on InfographicVQA using about half the image area. The analyses show that useful evidence depends on both its relevance to the question and the context available to the reader. \qcg\ makes these choices explicit before answer generation.
\end{minipage}
\vspace{9pt}
\begin{minipage}{\textwidth}
\centering
\includegraphics[width=\textwidth]{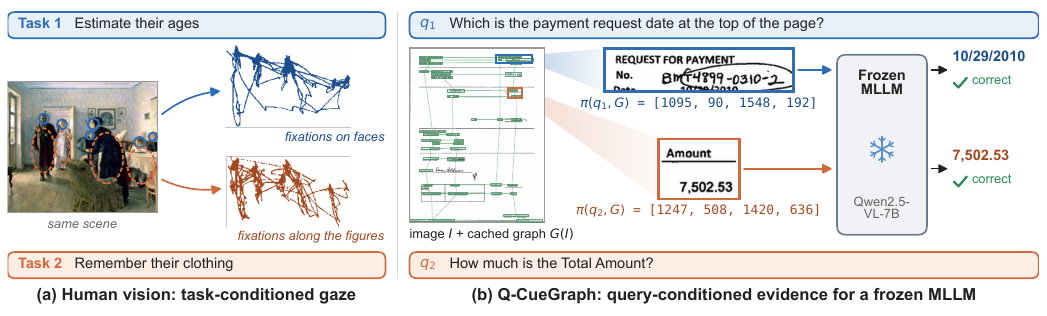}
\captionof{figure}{\textbf{Task-conditioned observation.} Left: different viewing instructions in Yarbus's study direct gaze toward different parts of the same painting~\citep{yarbus1967}. Historical traces are recolored; annotations on the painting illustrate their spatial emphasis. Right: questions about a date and an amount activate different evidence in the same cached document graph. \qcg\ returns image coordinates, and the frozen reader answers from the selected pixels.}
\label{fig:concept}
\end{minipage}
\vspace{10pt}
}]
\section{Introduction}
\label{sec:introduction}
A high-resolution image can contain all the information needed to answer a question while presenting that information at an ineffective scale. A document date may occupy a few characters in a header. An amount may be meaningful only beside its field label. A spatial relation requires both referenced objects to remain visible. Selecting an observation therefore involves two decisions: which content is relevant to the question, and which neighboring content is needed to interpret it.

Human vision offers a useful motivation for this distinction. In Yarbus's experiments, changing the viewing instruction changed the observer's gaze over the same painting~\citep{yarbus1967}. Models of goal-directed attention similarly condition visual search on a target~\citep{yang2020gaze,mondal2023gazeformer}. Figure~\ref{fig:concept} illustrates the corresponding question-answering problem: a page and its layout remain fixed, while questions about its date and amount require different observations. The relevant visual input is a function of the task as well as the image.

Selective visual processing has become an important component of multimodal reasoning. Guided search and learned tool use acquire local views during reasoning~\citep{wu2024vstar,zheng2026deepeyes}, while internal model signals can reveal useful regions~\citep{zhang2025vicrop,zhong2025focus}. Regional supervision can also improve full-image perception during training~\citep{yuan2026visionopd}. These approaches establish the value of closer inspection. An additional opportunity is to organize the evidence within an image so that each question can retrieve an appropriate observation before the answering model is called.

We propose \qcg, an external, query-conditioned evidence acquisition method for a frozen MLLM. For text-rich images, OCR lines form nodes and layout relations connect neighboring content. This graph is constructed once per image and reused across questions. Query terms activate anchors; directional expansion groups them with nearby values and context. The policy ranks candidate regions and composes the selected regions into one image window. For natural images, open-vocabulary detections provide query-conditioned visual nodes through the same selection and composition interface. The reader receives the resulting crop and the original question through its existing visual and text inputs.

This design separates locating relevant content from constructing evidence that the reader can use. A matched label is a good entry point, but its value may lie elsewhere in the same row. Two individually relevant objects may need to be observed together. Even a crop containing the answer string may lead to an incorrect response. To address this last distinction, we learn candidate utility from the frozen reader's crop-level answers. Training answers supply supervision for a small ranking model; the reader's parameters remain fixed. The ranking target is whether the observation supports a correct answer.

Our experiments follow these stages of evidence acquisition. Document ablations examine how query activation and layout expansion contribute to selection. Attribute and relation questions test the context preserved by composition. A controlled refinement study compares answer presence with reader answerability. Across six benchmarks, these analyses connect the benefit of selective observation to the spatial organization of the required evidence. They show strong gains for localized visual search, useful accuracy--area trade-offs for text-rich images, and the importance of retaining broader context for dispersed evidence. The resulting contribution is a concrete evidence policy whose representation, composition, and learning target can be studied independently of answer generation.

\section{Related Work}
\label{sec:related}
\subsection{Task-Conditioned Visual Observation}
Yarbus's experiments established the dependence of gaze on viewing instructions~\citep{yarbus1967}. Computational models learn target-conditioned search through inverse reinforcement learning~\citep{yang2020gaze} and transformer-based scanpath prediction~\citep{mondal2023gazeformer}. \qcg\ applies this task-conditioned perspective to evidence acquisition for question answering. Its target is an image observation that supports a downstream response, and its output specifies the pixels presented to the answering model.

\subsection{Visual Search, Cropping, and Retrieval}
V* combines an MLLM with guided visual search over high-resolution scenes~\citep{wu2024vstar}. DeepEyes learns visual tool use through reinforcement learning~\citep{zheng2026deepeyes}, and Visual CoT uses intermediate region annotations for visual reasoning~\citep{shao2024visualcot}. ZoomEye and DyFo explore image regions through hierarchical and Monte Carlo search~\citep{shen2025zoomeye,li2025dyfo}. These methods integrate observation into a sequence of reasoning steps. \qcg\ organizes candidate evidence through image structure and composes an observation before answer generation.

A second family obtains regions from the reader's internal signals. ViCrop uses attention and gradients~\citep{zhang2025vicrop}; FOCUS derives relevance maps from the key--value cache~\citep{zhong2025focus}; and self-distilled region predictors learn proposals from model-internal signals~\citep{shi2026sdrpn}. CropVLM learns an external cropping policy from task reward with a frozen reader~\citep{carvalho2025cropvlm}. \qcg\ instead builds document proposals from explicit lexical and layout relations, then uses supervised candidate ranking to refine their utility. This factorization makes the proposal structure and the ranking target accessible to separate analysis.

Within-image retrieval provides a related view of high-resolution perception. Retrieval-Augmented Perception retrieves and composes image crops~\citep{wang2025rap}, while VisRAG retrieves document images for generation~\citep{yu2025visrag}. \qcg\ operates within a supplied image. Its cached document graph supports repeated questions about the same page, with question-specific activation and expansion defining the candidate evidence.

\subsection{Learning from Regional Evidence}
Vision-OPD~\citep{yuan2026visionopd} transfers the benefit of regional perception into a full-image model through on-policy self-distillation. A crop-conditioned teacher supplies next-token distributions along trajectories sampled by a full-image-conditioned student, and the student learns by minimizing their token-level divergence. \qcg\ intervenes at the observation interface: it selects the visual input of a frozen reader at inference. Its optional learning stage uses training answers to supervise candidate-level utility. The two methods address the regional-to-global perception gap at different points---Vision-OPD improves the reader through parameter updates, whereas \qcg\ improves the evidence supplied to the reader through region selection and composition.

\section{Method}
\label{sec:method}
\subsection{Observation Policy and Reader Interface}
Let $I$ be an image, $q$ a question, and $f$ a frozen MLLM. \qcg\ forms a candidate set $P(I,q)$ from an evidence representation $S$. Text-rich images use a cached document graph $S=G_{\mathrm{doc}}(I)$; natural-image search uses a question-conditioned detection set $S=V_{\mathrm{vis}}(I,q)$. The policy ranks candidates and retains a set $C_K$ of at most $K$ regions. Its output is
\begin{equation}
 r_O=\pad\!\left(\bbox\!\left(\bigcup_{r\in C_K}r\right)\right),
 \qquad B=\frac{\area(r_O)}{\area(I)}.
 \label{eq:observation}
\end{equation}
Here $\bbox$ is the smallest enclosing axis-aligned rectangle, and padding is clipped to the image boundary. The observation is one rectangular window, including the space between its selected regions. The reader predicts
\begin{equation}
 \hat a=f\bigl(\crop(I,r_O),q\bigr).
 \label{eq:reader}
\end{equation}
The interface consists of crop pixels and the original question. Composition width $K$ and padding control observation extent; $B$ measures the resulting fraction of source-image area. Figure~\ref{fig:method} shows the selection and learning paths.

\fullfigure{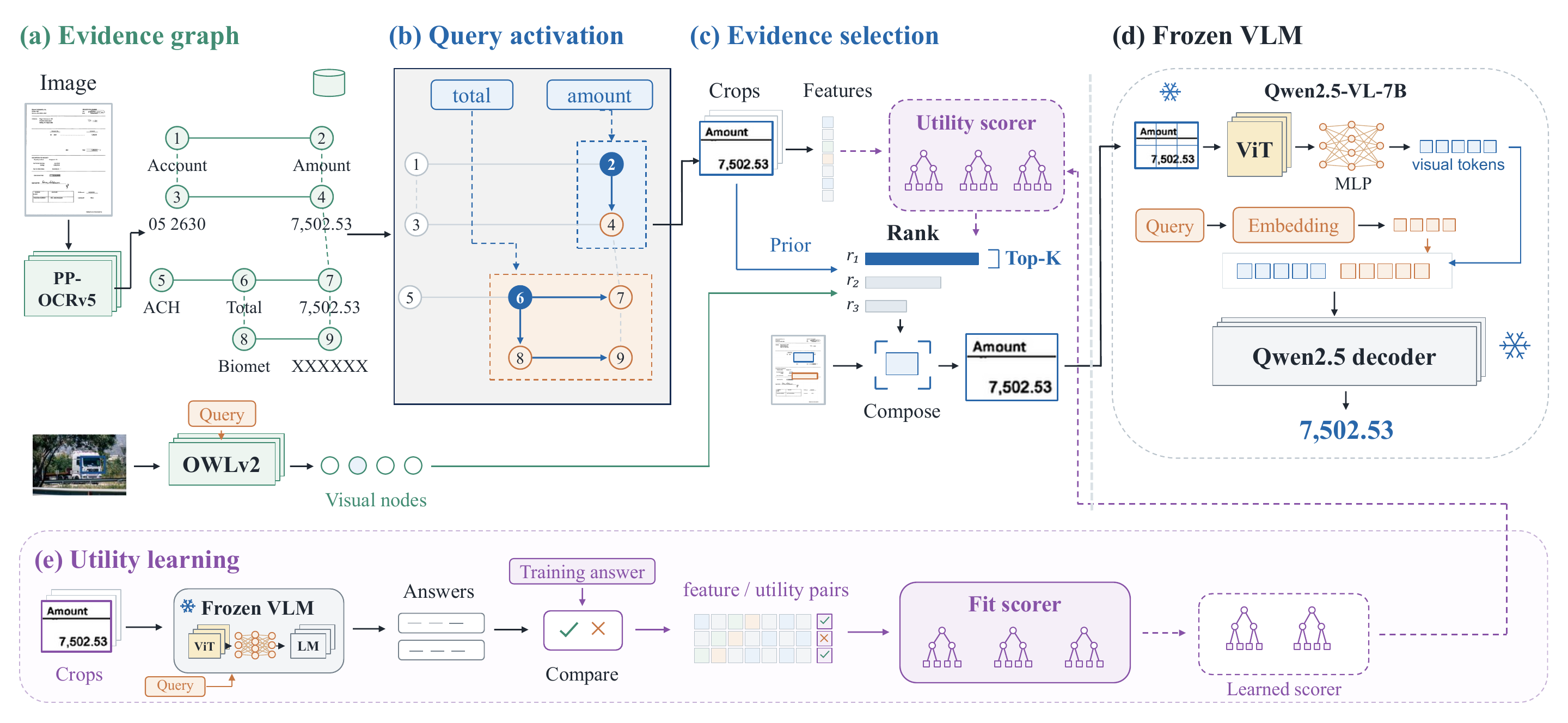}{\textbf{Q-CueGraph architecture.} (a) OCR lines and layout relations form a reusable document graph. (b) Query terms activate anchors, whose directional neighborhoods define candidate regions. (c) A structural prior or learned utility scorer ranks candidates, and top-$K$ composition produces one crop. Query-conditioned visual detections enter the same selection interface. (d) The crop and question are processed by the pretrained visual encoder, merger, and decoder of the frozen reader. (e) Offline, candidate responses are compared with training answers to supervise the utility scorer; the dashed path deploys the fitted scorer for ranking.}{fig:method}

\subsection{Reusable Document Structure}
PP-OCRv5 in PaddleOCR 3.0~\citep{cui2025paddle} extracts OCR lines. Each node $v$ stores its text, recognition confidence, and image-coordinate box $b_v$. Vertical and horizontal overlap define row and column relations. Directed adjacency identifies the nearest node to the right and below. The resulting graph $G_{\mathrm{doc}}(I)=(V,E)$ is cached for reuse across questions.

Page-specific labels and identifiers are more discriminative anchors than repeated boilerplate. A token $t$ receives the inverse line-frequency weight
\begin{equation}
 w_t=\log\frac{N+1}{n_t+0.5},
 \label{eq:rarity}
\end{equation}
where $N=|V|$ is the number of OCR lines and $n_t$ counts lines containing $t$. These weights are cached with the graph. After stopword removal, question content words are fuzzily matched to OCR text at a similarity threshold of 0.82. Denote their rarity-weighted lexical score by $m(v,q)$. The anchor score is
\begin{equation}
 a(v,q)=
 \begin{cases}
 m(v,q)+\lambda_c\min\{\deg(v),5\}, & m(v,q)>0,\\
 0, & \text{otherwise},
 \end{cases}
 \label{eq:anchor}
\end{equation}
with $\lambda_c=0.02$ and node degree $\deg(v)$ under the layout relations. The centrality term favors connected nodes among positive lexical matches.

For each positive anchor, expansion combines its own box, its rightward row segment, and the row associated with its below neighbor. The enclosing region becomes a candidate, scored by the anchor. Candidates with nearly coincident centers are suppressed. Thus, a question can locate a field name while layout brings its value into the same observation.

The ungated policy retains at least one candidate. When every anchor score is zero, stable sorting preserves the OCR output order, and the first OCR line is expanded into a query-independent structural fallback. An image with no OCR lines uses the full view. An anchor-gated policy instead uses the full view whenever all anchor scores are zero. Appendix~\ref{app:fallback} gives the corresponding decision rule.

\subsection{Visual Nodes and Contextual Composition}
For natural images, OWLv2~\citep{minderer2023owl} is queried with question content words and bigrams. Its scored detections form $V_{\mathrm{vis}}(I,q)$, a per-question node set without structural edges. Each detection directly supplies a candidate, and detector confidence supplies its ranking score.

Both branches apply top-$K$ selection and the composition in Eq.~\eqref{eq:observation}. The enclosing window preserves relative image positions, allowing two referents to be read together in a spatial-relation question. Padding restores context around an OCR line or detected object. Document padding, scene-context padding, and the wide-union infographic configuration specify different observation extents; their constants appear in Appendix~\ref{app:implementation}. The visual branch uses the full image when detection returns an empty set.

\subsection{Learning Candidate Answerability}
\label{sec:utility}
The structural prior estimates relevance from the question and image layout. We refine it by learning which candidate observations support correct responses from the reader. On a training partition, each candidate $r_j$ is rendered and read with its question. Given training answer $a^*$, its utility label is
\begin{equation}
 u_j=\mathbf{1}\!\left[\ANLS\!\left(f(\crop(I,r_j),q),a^*\right)\geq 0.5\right],
 \label{eq:utility}
\end{equation}
where ANLS denotes average normalized Levenshtein similarity~\citep{biten2019stvqa}. A gradient-boosted scorer $s_\phi(\psi_j)$ is fitted to these labels. Its seven candidate features are anchor score, relative rank, area, OCR count, question-token coverage, candidate index, and pool size. At inference, it reranks the candidate pool using these features before top-$K$ composition.

Training answers supervise the scorer, while the reader remains frozen. The document branch supports both the structural prior and learned ranking; the natural-image experiments use detector scores. For a controlled target comparison, a second document scorer uses the same candidate features but learns whether the crop's OCR contains the reference answer. This localization target measures answer presence, whereas $u_j$ measures answerability under the fixed reader and evaluation protocol.

\subsection{Computation}
OCR and document-relation construction are one-time image costs. On 300 cached DocVQA queries, graph activation and candidate selection take 1.4\,ms on average on one CPU core. OCR construction takes approximately 0.1\,s per image on one GPU in the measured setup. Natural-image search requires one detector pass per question. The selected window is then rendered and passed to the reader for answer generation. Timing quantiles, operation costs, and rendered-pixel measurements appear in Appendix~\ref{app:implementation}.

\section{Results}
\label{sec:results}
\subsection{Evaluation Protocol}
All comparisons use a frozen Qwen2.5-VL-7B reader~\citep{bai2025qwen} through Ollama, with temperature 0 and at most 32 generated tokens. Prompts and answer scoring are fixed within each benchmark. We evaluate DocVQA~\citep{mathew2021docvqa}, InfographicVQA~\citep{mathew2022infovqa}, TextVQA~\citep{singh2019textvqa}, OCRBench~\citep{liu2024ocrbench}, ChartQA~\citep{masry2022chartqa}, and \vstar~\citep{wu2024vstar}. The tasks span document fields, dispersed layouts, scene text, charts, and localized visual attributes or relations.

Full-benchmark comparisons evaluate the structural or detector prior; learned ranking is studied separately. TextVQA and InfographicVQA exclude the 150 and 20 examples used for adapter calibration. DocVQA tables report all 5,349 validation questions, with an additional analysis excluding the first 500 used to develop directional expansion. Utility refinement uses a contiguous 1,494/498/499 train/validation/test partition of 2,491 DocVQA validation questions. Appendix~\ref{app:protocol} gives the split construction and calibration analysis.

High-resolution full views have a maximum rendered side of 1,536 pixels, low-resolution full views 512 pixels, and crops 1,280 pixels. Baselines comprise generic crops, query-only and graph-only ablations, displaced anti-regions, and shuffled-question policies. We report source-image area alongside answer scores. Appendix~\ref{app:grids} contains all 79 conditions and the task-specific configurations.

\subsection{Questions Select Anchors; Layout Supplies Context}
\label{sec:query_results}
Document questions often name a key while asking for its neighboring value. Table~\ref{tab:docmechanism} examines the contribution of query and layout information to this selection problem. Question-only matching reaches 0.371 ANLS, and graph-only selection reaches 0.130. Their combination in \qcg\ top-1 reaches 0.508 at 5.7\% of source-image area. The query provides an entry point into the page; expansion turns that entry point into a region containing related content.

\begin{table}[t]
\caption{\textbf{Query and structure on DocVQA.} ANLS on 5,349 validation questions; $B$ is mean source-image area. Observation and fallback definitions appear in Appendix~\ref{app:controls}.}
\label{tab:docmechanism}
\centering\smalltable
\begin{tabular}{@{}lrr@{}}
\toprule
Observation & ANLS & $B$\\
\midrule
Full image & 0.903 & 1.000\\
Center crop & 0.546 & 0.431\\
Question only & 0.371 & 0.186\\
Graph only & 0.130 & 0.055\\
\midrule
\qcg\ top-1 & 0.508 & 0.057\\
\qcg\ top-2 & 0.650 & 0.162\\
\qcg\ top-4 & 0.712 & 0.249\\
\midrule
Shuffled question & 0.224 & 0.062\\
Anti-region & 0.077 & 0.046\\
\bottomrule
\end{tabular}
\end{table}

Running the policy with a shuffled question lowers ANLS to 0.224, while reading a displaced region lowers it to 0.077. The selected content therefore matters beyond a generic preference for a dense part of the page. Same-page comparisons reveal the corresponding change in coordinates: across 11,857 pairs of questions sharing a page, 65.7\% of top-1 window pairs have intersection-over-union below 0.1. One cached representation supports substantially different observations as the question changes (Appendix~\ref{app:samepage}). Excluding the 500 development questions changes the top-1, top-2, and top-4 ANLS values by less than 0.003 (Table~\ref{tab:calibration}).

\subsection{Composition Preserves the Evidence Relation}
\label{sec:extent_results}
A useful observation must retain the relation that the question asks the reader to recover. On \vstar, this requirement distinguishes single-object attributes from two-object spatial relations (Table~\ref{tab:composition}). Top-1 and top-2 have similar attribute accuracy, whereas adding the second region raises relation accuracy from 0.737 to 0.855. The qualitative cases show both referents entering the composed window. The benefit of additional area thus depends on what that area contributes to the question.

\begin{table}[t]
\caption{\textbf{Composition and question type on \vstar.} Top-2 composition primarily benefits relative-position questions. All three policies use the same frozen crop reader.}
\label{tab:composition}
\centering\smalltable
\begin{tabular}{@{}lrrrr@{}}
\toprule
Question type & $n$ & Self-zoom & Top-1 & Top-2\\
\midrule
Attribute & 115 & 0.878 & 0.826 & 0.817\\
Relation & 76 & 0.684 & 0.737 & 0.855\\
\bottomrule
\end{tabular}
\end{table}

The same principle has a different geometry in text-rich images. On DocVQA, top-2 composition exceeds a center crop while using substantially less of the page; top-4 reaches 79\% of full-image ANLS at 24.9\% area. Infographics distribute related evidence across wider layouts. Their narrow top-1 window reaches 0.282 ANLS, while a wide top-2 union reaches 0.686 at 50.5\% area, retaining 92\% of the full-image score.

For scene text, the surrounding object can determine what an OCR span refers to. On anchored TextVQA questions, accuracy increases from 0.223 for the bare OCR line to 0.420 after structural expansion and 0.528 with scene-context padding. Together, the document, infographic, and scene-text results identify an important role for the anchor: it provides access to relevant context, not just to the characters that match the question.

\subsection{Learning to Rank Reader-Usable Evidence}
\label{sec:utility_results}
Answer presence and answerability provide different supervision for observation selection. Among 19,221 candidate crops whose OCR contains the reference answer, 2,699 (14\%) receive an incorrect crop-only answer score. A location containing the answer is therefore a useful starting point, but the observation's utility also depends on the response it supports.

\begin{figure}[t]
\centering
\includegraphics[width=\columnwidth]{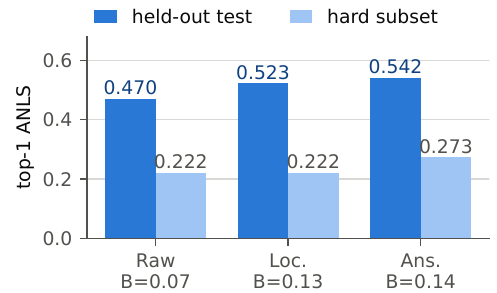}
\caption{\textbf{Refinement targets.} Top-1 ANLS on the 499-question test partition and its full-image-error subset. Raw: structural prior; Loc.: localization supervision; Ans.: answerability supervision. Mean area $B$ is shown for each target.}
\label{fig:utility}
\end{figure}

Figure~\ref{fig:utility} compares ranking targets on the 499-question test partition. Answerability supervision reaches 0.542 ANLS, compared with 0.523 for answer-string localization, at mean areas of 0.140 and 0.129. The difference increases on questions that full-image reading answers incorrectly: 0.273 versus 0.222. Reader feedback directs ranking toward observations that better support answering, with the largest gain on this harder subset.

The same scorer transfers from DocVQA to InfographicVQA: wide-union ANLS increases from 0.686 to 0.695, and mean area changes from 0.505 to 0.542. This transfer applies the learned ranking directly to a different document layout. Appendix~\ref{app:refinement} reports the complete target comparison and the separate candidate-selection analysis.

A structure-only probe clarifies the information used at each stage. On 600 DocVQA questions, OCR/layout serialization reaches 0.684 ANLS and selected pixel crops reach 0.736. Removing lexical content or retaining only matched anchors sharply reduces the serialization result (Appendix~\ref{app:structure}). Recognized text and its surrounding structure guide selection; the resulting image window supplies the visual input for answering. This division explains why the policy can use a compact structural representation while retaining the pretrained reader's visual capabilities.

\subsection{Matching the Observation to the Evidence}
\label{sec:regime_results}
\begin{figure}[t]
\centering
\includegraphics[width=\columnwidth]{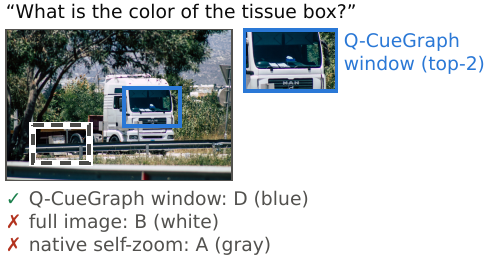}
\caption{\textbf{A change of observation changes the answer.} Recorded outputs for the same question and frozen reader. The \qcg\ window supports the correct tissue-box color, while the full view and native self-zoom produce different, incorrect answers.}
\label{fig:vstar_example}
\end{figure}

\vstar\ provides the clearest improvement over full-image reading. Top-2 observations reach 0.832 accuracy versus 0.696 for the full view, using 19.1\% of source-image area and approximately 25\% of the rendered input pixels. The paired improvement is $+0.136$ (95\% confidence interval $[+0.068,+0.204]$, exact McNemar $p=0.0002$), comprising 36 corrected answers and 10 regressions. Concentrating the visual input on question-relevant content helps recover small details in large scenes (Figure~\ref{fig:vstar_example}).

Native self-zoom attains 0.801 accuracy at 8.6\% mean area, compared with 0.832 at 19.1\% for \qcg\ top-2 (paired $p=0.43$). Their errors reveal complementary evidence choices: 23 questions are answered correctly only by \qcg, and 17 only by self-zoom. The hindsight either-correct score is 0.921. Their strengths also vary by question type: self-zoom is stronger on attributes, while \qcg\ top-2 is stronger on relations (Table~\ref{tab:composition}). This pattern connects coordinate selection with the contextual requirements of the answer.

Anchor availability explains the role of fallback on scene text. Of 4,850 TextVQA questions, 3,613 have no positive lexical anchor. On this subset, ungated and shuffled-question windows coincide in 94\% of cases, reflecting the same structural default. Anchor gating reserves local observations for positive matches and uses the full view otherwise. With context padding, it reaches 0.672 accuracy at 0.808 mean area; the corresponding OCRBench configuration reaches 0.648 at 0.723 area. Appendix~\ref{app:fallback_audit} gives the no-anchor counts and control behavior.

ChartQA illustrates the need for broader evidence: top-1 and shuffled-question scores are close (0.246 and 0.238), while full-image inference reaches 0.785. A lexical anchor can be present even when the answer requires multiple axes or series. Across the benchmarks, observation quality depends on the concentration of the evidence, the question's ability to identify its location, and the detail retained at the reader's input scale. The operating points in Appendix~\ref{app:regimes} organize this relationship from compact visual search to wide contextual observation.

\subsection{Scope and Limitations}
\label{sec:scope}
\qcg\ uses a fixed candidate extractor, explicit document relations, and a single enclosing window. Evidence omitted by extraction is unavailable to ranking, and distant regions can consume substantial area when enclosed together. Composition width and task adapters are specified configurations rather than a learned cross-task controller. The operating-regime analysis is retrospective, and the either-correct result assumes hindsight selection. Adaptive composition and iterative tool integration remain extensions of the present method.

All reader experiments use Qwen2.5-VL-7B and one deterministic run per condition. Utility refinement partitions questions in manifest order, without grouping by page; its held-out results therefore do not establish page-disjoint or cross-reader generalization. The exact number of shared pages has not been established from image identities. Development-slice exclusion is reported for DocVQA, and TextVQA and InfographicVQA exclude their adapter-calibration examples. Evaluation across reader architectures would be needed to establish a general-purpose visual layer.

Answerability is defined by a fixed prompt, decoder, and metric, and includes response-format compliance as well as visual understanding. Changes in ranking can also change area, as in the localization/answerability and transfer comparisons. Cropping jointly affects scale, context, and distractors; the experiments do not isolate their individual causal effects. Finally, source-image area characterizes observation extent, rather than a hard optimization constraint or an end-to-end speedup. These distinctions define the interpretation of the reported gains.

\balance
\section{Conclusion}
We presented \qcg, a query-conditioned evidence acquisition method that addresses where a frozen MLLM should look before answering. It organizes OCR and layout into a reusable document graph, or instantiates question-conditioned visual nodes for natural images, and turns selected regions into a contextual observation. An answer-supervised utility scorer further improves candidate ranking while preserving the reader's pretrained parameters.

Across six benchmarks, the experiments establish the distinct roles of selecting relevant content, preserving its context, and learning its utility for the reader. \qcg\ improves fine-grained visual search and provides useful accuracy--area trade-offs for document and infographic understanding. The central insight is that an effective observation must capture both the evidence named by a question and the context needed to interpret it. By making this decision explicit, \qcg\ provides a practical method for improving the visual evidence available to multimodal reasoning.

\clearpage
\nobalance
{\small\setlength{\bibsep}{1pt}\bibliographystyle{unsrt}\bibliography{references}}
\clearpage
\appendix
\setcounter{figure}{0}
\setcounter{table}{0}
\renewcommand{\thefigure}{A\arabic{figure}}
\renewcommand{\thetable}{A\arabic{table}}
\renewcommand{\theHfigure}{appendix.\arabic{figure}}
\renewcommand{\theHtable}{appendix.\arabic{table}}
\section{Implementation Details}
\label{app:implementation}
\subsection{Graph Operations and Candidate Geometry}
The document representation contains one node per PP-OCRv5 line. Same-row relations use vertical overlap; same-column relations use horizontal overlap. Nearest-right and nearest-below adjacency determine the neighborhoods used to form regions. Question tokens are matched at similarity threshold 0.82. The positive-match centrality contribution is weighted by 0.02 and capped at degree five. Page-local token rarity uses the number of OCR lines containing each token.

Expansion combines the anchor box, its rightward row segment, and the row associated with its below neighbor. Candidate centers within 40 horizontal and 25 vertical pixels are treated as near duplicates. The selected regions are enclosed in one rectangle, padded, and clipped to the original image. The area therefore includes gaps between separated candidates and counts overlapping content once. Table~\ref{tab:settings} lists the configuration constants.

\begin{table}[t]
\caption{\textbf{Observation settings.} Percentages specify padding or minimum-side settings.}
\label{tab:settings}
\centering\smalltable
\begin{tabular}{@{}p{0.43\columnwidth}p{0.49\columnwidth}@{}}
\toprule
Component & Setting\\
\midrule
Fuzzy-match threshold & 0.82\\
Centrality & Weight 0.02; degree cap 5\\
Near-duplicate centers & 40 px horizontal; 25 px vertical\\
Document padding & 4\% plus 8 px\\
Scene-context padding & 50\%; minimum side 35\% of image\\
Infographic wide union & Minimum side 50\% of image\\
Visual top-1 & 100\% padding; minimum side 25\%\\
Visual top-2 & Union, then 50\% padding\\
Crop render & Maximum side 1,280 px\\
Full-image render & Maximum side 1,536 px\\
Low-resolution render & Maximum side 512 px\\
\bottomrule
\end{tabular}
\end{table}

\subsection{Structural Fallback and Anchor Gating}
\label{app:fallback}
Document nodes are sorted by descending anchor score, with ties preserving their order in the OCR output. Candidate generation uses a minimum count of one: a zero-score node stops expansion only after at least one candidate has been formed. If all scores are zero, this rule expands the first OCR line into a structural default. An empty OCR output produces an empty candidate pool and triggers the full-image fallback. The gated variant additionally uses the full image whenever no positive anchor exists. Algorithm~\ref{alg:document} states these branches explicitly.

\begin{algorithm}[t]
\caption{Document observation with structural fallback}
\label{alg:document}
\small
\begin{algorithmic}[1]
\Require Image $I$, question $q$, cached OCR nodes $V$, layout relations $E$, width $K$, gate flag $g$
\Ensure Observation window $r_O$
\If{$V=\varnothing$}
  \State \Return full-image window
\EndIf
\State Compute anchor scores $a(v,q)$ by Eq.~\eqref{eq:anchor}
\State $h\gets [\max_{v\in V}a(v,q)>0]$
\If{$g$ and not $h$}
  \State \Return full-image window
\EndIf
\If{$h$}
  \State Stable-sort $V$ by decreasing $a(v,q)$
  \State Expand positive anchors and suppress duplicates
  \State $P\gets$ resulting scored candidate pool
\Else
  \State $v_0\gets$ first node in OCR output order
  \State $P\gets\{(\operatorname{expand}(v_0,E),0)\}$
\EndIf
\State Rank $P$ by its prior or the optional utility scorer
\State $C_K\gets$ highest-ranked $K$ candidates in $P$
\State \Return $\pad(\bbox(\bigcup_{r\in C_K}r))$
\end{algorithmic}
\end{algorithm}

The question-only ablation has its own fallback: lexical matching returns no region when no anchor is positive, so that condition reads the full image. This differs from the ungated structural default and is included in the reported observation areas.

\subsection{Candidate Features and Refinement}
The seven features in Section~\ref{sec:utility} encode the prior, candidate extent, lexical support, and proposal-set position. Training pairs these features with the frozen reader's answer score for each rendered candidate to fit a gradient-boosted utility scorer.

The localization comparator uses the same features and candidates, replacing answerability with answer-string presence in OCR. Both supervise individual crops; top-$K$ composition follows ranking.

\subsection{Graph Operation Costs and Measured Time}
Let $N$ be the number of OCR nodes, $\bar t$ the mean tokens per node, $N_+$ the number of positive anchors, $|\tilde q|$ the number of question content tokens, and $|P|$ the candidate count. Lexical matching, query-time centrality, and expansion/deduplication contribute
\begin{equation}
 O(|\tilde q|N\bar t)+O(N_+N)+O(|P|N),
 \label{eq:complexity}
\end{equation}
respectively. The implementation computes the degree contribution at query time; node and candidate scores are also sorted for selection.

For 300 queries on cached DocVQA graphs, one-core CPU time is 1.4\,ms on average, 0.8\,ms at the median, and 5.0\,ms at the 95th percentile. One-time OCR construction takes approximately 0.1\,s per image on one GPU in the measured setup. Natural-image search adds one detector pass per question.

Relative to the high-resolution full view, mean rendered-input pixel fractions are approximately 0.25 for \vstar\ top-2, 0.24 for DocVQA top-2, and 0.72 for the InfographicVQA wide union. Their source-area fractions are approximately 0.19, 0.16, and 0.51. Each observation is resized independently before reading.

\section{Evaluation and Calibration Protocols}
\label{app:protocol}
\subsection{Evaluation Sets}
Table~\ref{tab:datasets} gives the evaluation sets and metrics. Conditions share the reader, prompt, answer cleaning, and scoring code within each benchmark. Ollama uses temperature 0 and a 32-token output limit for one deterministic run. Reference boxes supply post-hoc \vstar\ overlays.

\begin{table}[t]
\caption{\textbf{Evaluation sets and metrics.} The TextVQA and InfographicVQA counts exclude adapter-calibration examples.}
\label{tab:datasets}
\centering\smalltable
\begin{tabular}{@{}lrl@{}}
\toprule
Benchmark & Questions & Primary metric\\
\midrule
DocVQA validation & 5,349 & ANLS\\
InfographicVQA validation & 2,781 & ANLS\\
TextVQA validation & 4,850 & VQA accuracy\\
OCRBench test & 1,000 & Substring hit\\
ChartQA test & 2,500 & Relaxed accuracy\\
\vstar & 191 & MC accuracy\\
\bottomrule
\end{tabular}
\end{table}

\subsection{Configuration Selection and Exclusion Analysis}
The document grid $K\in\{1,2,4\}$ was declared before the full runs. Visual top-1/top-2 composition and padding were fixed when the \vstar\ runner was written. The final directional expansion rule was developed using DocVQA-500, the first 500 validation questions. Tables~\ref{tab:docmechanism} and \ref{tab:grid_doc} use the complete 5,349-question split. Table~\ref{tab:calibration} additionally reports the 4,849 questions remaining after exclusion. The largest ANLS change is 0.0026; mean areas agree at the precision reported in the complete grid.

\begin{table}[t]
\caption{\textbf{DocVQA development-slice exclusion.} ANLS is shown to four decimals to expose the changes. The excluded set is the first 500 validation questions.}
\label{tab:calibration}
\centering\smalltable
\begin{tabular}{@{}lrrr@{}}
\toprule
Policy & All 5,349 & Exclude 500 & Change\\
\midrule
Top-1 & 0.5075 & 0.5076 & $+0.0001$\\
Top-2 & 0.6499 & 0.6489 & $-0.0010$\\
Top-4 & 0.7120 & 0.7094 & $-0.0026$\\
\bottomrule
\end{tabular}
\end{table}

TextVQA gating and scene-context padding were selected using the first 150 validation examples. The InfographicVQA wide union was selected after a 20-example calibration run. Results for these benchmarks exclude those calibration slices. The complete grids retain the top-1 policy and all task-adapted configurations. The best-variant display in Figure~\ref{fig:regimes} is a post-hoc envelope over the unrefined configurations.

\subsection{Utility-Refinement Partitions}
\label{app:refinement_split}
The refinement set contains 2,491 DocVQA validation questions. A contiguous split in manifest order assigns the first 60\% to training, the next 20\% to validation, and the final 20\% to testing, giving 1,494/498/499 questions. Candidate labels and scorer fitting use training questions. The target comparison uses the 499-question test set, recorded at original manifest positions 2001--2499, separate from DocVQA-500. The split unit is the question; page identity is not used to form the partitions. Appendix~\ref{app:refinement} also retains the separate 1,605-question selector analysis.

\subsection{Reader and Zoom Prompts}

The document and infographic reader prompt is:
\begin{quote}\small\ttfamily\raggedright
Answer the question using a single word or short phrase taken directly from the document. Question: \{question\}\\
Answer:
\end{quote}
TextVQA uses:
\begin{quote}\small\ttfamily\raggedright
Answer the question using a single word or short phrase based on the text in the image. Question: \{question\}\\
Answer:
\end{quote}
OCRBench uses:
\begin{quote}\small\ttfamily\raggedright
\{question\}\\
Answer with the exact text or a single word/short phrase.\\
Answer:
\end{quote}
ChartQA uses:
\begin{quote}\small\ttfamily\raggedright
Answer the question about the chart with a single number or a short answer. Question: \{question\}\\
Answer:
\end{quote}
\vstar\ passes the benchmark question and choices verbatim, followed by \texttt{Answer:}. Native self-zoom first prompts the same reader with the full image:
\begin{quote}\small\ttfamily\raggedright
You can zoom into one region of this image to answer a question later. Question: \{question\}\\
Reply ONLY with the bounding box of the region to zoom into, as [x0, y0, x1, y1] pixel coordinates of THIS image. No other text.
\end{quote}

The shared renderer processes valid native windows; invalid coordinates use the full view. The reader then receives the benchmark question followed by \texttt{Answer:}.

\subsection{Controls}
\label{app:controls}
Question-only selection uses lexical matches, with full-image fallback on no match; graph-only selection removes the question. Anti-regions displace the observation, and shuffled controls use another example's question. Generic crops use the center, a random region, or OCR density. Area includes the final window and any fallback: the \vstar\ shuffled control averages 0.397, versus 0.134 for top-1.

\onecolumn
\section{Complete Condition Grids}
\label{app:grids}
Tables~\ref{tab:grid_doc}--\ref{tab:grid_vstar} report all 79 conditions. $B$ is mean source-image area. Top-$K$ denotes \qcg; ``refined'' rows use learned ranking. Gated conditions use full-view fallback on no-anchor questions (74.5\% of TextVQA and 62.7\% of OCRBench). Each table includes the full-image reference.

\noindent\begin{minipage}[t]{0.485\textwidth}
\vspace{0pt}
\captionsetup{font=footnotesize,skip=4pt}
\captionof{table}{\textbf{DocVQA.} $n=5,349$. Validation; ANLS and exact match.}
\label{tab:grid_doc}
\centering\fontsize{8.5}{9.6}\selectfont\setlength{\tabcolsep}{4pt}\renewcommand{\arraystretch}{1.01}
\begin{tabular}{@{}lrrr@{}}
\toprule
Observation & ANLS & EM & $B$\\
\midrule
Full image & 0.903 & 0.814 & 1.000\\
Low-resolution full view & 0.869 & 0.744 & 1.000\\
Center crop & 0.546 & 0.442 & 0.431\\
Random crop & 0.202 & 0.132 & 0.146\\
OCR-density crop & 0.317 & 0.254 & 0.155\\
Question only & 0.371 & 0.311 & 0.186\\
Graph only & 0.130 & 0.090 & 0.055\\
\addlinespace[2pt]
Top-1 & 0.508 & 0.435 & 0.057\\
Top-2 & 0.650 & 0.572 & 0.162\\
Top-4 & 0.712 & 0.635 & 0.249\\
\addlinespace[2pt]
Anti-region & 0.077 & 0.046 & 0.046\\
Shuffled question & 0.224 & 0.172 & 0.062\\
\bottomrule
\end{tabular}
\end{minipage}\hfill
\begin{minipage}[t]{0.485\textwidth}
\vspace{0pt}
\captionsetup{font=footnotesize,skip=4pt}
\captionof{table}{\textbf{InfographicVQA.} $n=2,781$. Validation, excluding 20 calibration examples.}
\label{tab:grid_info}
\centering\fontsize{8.5}{9.6}\selectfont\setlength{\tabcolsep}{4pt}\renewcommand{\arraystretch}{1.01}
\begin{tabular}{@{}lrr@{}}
\toprule
Observation & ANLS & $B$\\
\midrule
Full image & 0.743 & 1.000\\
Low-resolution full view & 0.533 & 1.000\\
Center crop & 0.499 & 0.437\\
Random crop & 0.298 & 0.149\\
OCR-density crop & 0.310 & 0.173\\
Question only & 0.185 & 0.028\\
Graph only & 0.170 & 0.064\\
\addlinespace[2pt]
Top-1 & 0.282 & 0.045\\
Top-2 & 0.530 & 0.251\\
Top-4 & 0.670 & 0.481\\
Gated top-1 & 0.292 & 0.063\\
Gated, padded top-1 & 0.561 & 0.255\\
Wide top-2 & 0.686 & 0.505\\
Refined top-1 & 0.301 & 0.072\\
Refined wide top-2 & 0.695 & 0.542\\
\addlinespace[2pt]
Anti-region & 0.164 & 0.055\\
Shuffled question & 0.188 & 0.050\\
\bottomrule
\end{tabular}
\end{minipage}
\par\vspace{7pt}
\noindent\begin{minipage}[t]{0.485\textwidth}
\vspace{0pt}
\captionsetup{font=footnotesize,skip=4pt}
\captionof{table}{\textbf{TextVQA.} $n=4,850$. Validation, excluding 150 calibration examples.}
\label{tab:grid_text}
\centering\fontsize{8.5}{9.6}\selectfont\setlength{\tabcolsep}{4pt}\renewcommand{\arraystretch}{1.01}
\begin{tabular}{@{}lrr@{}}
\toprule
Observation & Accuracy & $B$\\
\midrule
Full image & 0.708 & 1.000\\
Low-resolution full view & 0.694 & 1.000\\
Center crop & 0.530 & 0.444\\
Random crop & 0.199 & 0.153\\
OCR-density crop & 0.354 & 0.099\\
Question only & 0.594 & 0.754\\
Graph only & 0.373 & 0.136\\
\addlinespace[2pt]
Top-1 & 0.384 & 0.126\\
Top-2 & 0.392 & 0.139\\
Top-4 & 0.396 & 0.144\\
Gated top-1 & 0.645 & 0.774\\
Gated, padded top-1 & 0.672 & 0.808\\
\addlinespace[2pt]
Anti-region & 0.365 & 0.128\\
Shuffled question & 0.357 & 0.129\\
\bottomrule
\end{tabular}
\end{minipage}\hfill
\begin{minipage}[t]{0.485\textwidth}
\vspace{0pt}
\captionsetup{font=footnotesize,skip=4pt}
\captionof{table}{\textbf{OCRBench.} $n=1,000$. Substring-hit accuracy and auxiliary ANLS.}
\label{tab:grid_ocr}
\centering\fontsize{8.5}{9.6}\selectfont\setlength{\tabcolsep}{4pt}\renewcommand{\arraystretch}{1.01}
\begin{tabular}{@{}lrrr@{}}
\toprule
Observation & Accuracy & ANLS & $B$\\
\midrule
Full image & 0.786 & 0.871 & 1.000\\
Low-resolution full view & 0.760 & 0.856 & 1.000\\
Center crop & 0.388 & 0.596 & 0.537\\
Random crop & 0.098 & 0.176 & 0.207\\
OCR-density crop & 0.458 & 0.562 & 0.471\\
Question only & 0.520 & 0.619 & 0.637\\
Graph only & 0.442 & 0.548 & 0.486\\
\addlinespace[2pt]
Top-1 & 0.523 & 0.615 & 0.473\\
Top-2 & 0.584 & 0.679 & 0.531\\
Top-4 & 0.619 & 0.711 & 0.569\\
Gated top-1 & 0.609 & 0.700 & 0.663\\
Gated, padded top-1 & 0.648 & 0.745 & 0.723\\
\addlinespace[2pt]
Anti-region & 0.401 & 0.510 & 0.473\\
Shuffled question & 0.446 & 0.538 & 0.473\\
\bottomrule
\end{tabular}
\end{minipage}
\par\vspace{7pt}
\noindent\begin{minipage}[t]{0.485\textwidth}
\vspace{0pt}
\captionsetup{font=footnotesize,skip=4pt}
\captionof{table}{\textbf{ChartQA.} $n=2,500$. Test; relaxed answer accuracy.}
\label{tab:grid_chart}
\centering\fontsize{8.5}{9.6}\selectfont\setlength{\tabcolsep}{4pt}\renewcommand{\arraystretch}{1.01}
\begin{tabular}{@{}lrr@{}}
\toprule
Observation & Accuracy & $B$\\
\midrule
Full image & 0.785 & 1.000\\
Low-resolution full view & 0.751 & 1.000\\
Center crop & 0.445 & 0.452\\
Random crop & 0.181 & 0.158\\
OCR-density crop & 0.198 & 0.102\\
Question only & 0.283 & 0.238\\
Graph only & 0.219 & 0.155\\
\addlinespace[2pt]
Top-1 & 0.246 & 0.101\\
Top-2 & 0.316 & 0.226\\
Top-4 & 0.336 & 0.271\\
Gated top-1 & 0.371 & 0.304\\
Gated, padded top-1 & 0.395 & 0.404\\
\addlinespace[2pt]
Anti-region & 0.144 & 0.072\\
Shuffled question & 0.238 & 0.097\\
\bottomrule
\end{tabular}
\end{minipage}\hfill
\begin{minipage}[t]{0.485\textwidth}
\vspace{0pt}
\captionsetup{font=footnotesize,skip=4pt}
\captionof{table}{\textbf{V*Bench.} $n=191$. Test; multiple-choice accuracy.}
\label{tab:grid_vstar}
\centering\fontsize{8.5}{9.6}\selectfont\setlength{\tabcolsep}{4pt}\renewcommand{\arraystretch}{1.01}
\begin{tabular}{@{}lrr@{}}
\toprule
Observation & Accuracy & $B$\\
\midrule
Full image & 0.696 & 1.000\\
Low-resolution full view & 0.660 & 1.000\\
Center crop & 0.524 & 0.360\\
Random crop & 0.435 & 0.122\\
\addlinespace[2pt]
Visual top-1 & 0.791 & 0.134\\
Visual top-2 & 0.832 & 0.191\\
\addlinespace[2pt]
Anti-region & 0.393 & 0.134\\
Shuffled question & 0.618 & 0.397\\
\bottomrule
\end{tabular}
\end{minipage}

\twocolumn
\section{Supporting Analyses}
\subsection{Candidate Answerability and Reference-Guided Selection}
\label{app:refinement}
Table~\ref{tab:refinement} reports the candidate-ranking targets underlying Figure~\ref{fig:utility}. The hard subset contains questions answered incorrectly from the full image under the same reader protocol. Each row reports the mean area of the windows selected by that target.

\begin{table}[t]
\caption{\textbf{Candidate-ranking targets.} DocVQA internal test partition ($n=499$). Hard denotes the full-image-error subset. Scores are ANLS.}
\label{tab:refinement}
\centering\smalltable
\begin{tabular}{@{}lrrr@{}}
\toprule
Ranking target & All & $B$ & Hard\\
\midrule
Structural prior & 0.470 & 0.068 & 0.222\\
Localization & 0.523 & 0.129 & 0.222\\
Answerability & 0.542 & 0.140 & 0.273\\
\bottomrule
\end{tabular}
\end{table}

The candidate-pool diagnostic covers nine candidates per example across 5,349 questions. Of the 19,221 candidates whose OCR contains the reference answer, 2,699 receive an incorrect crop-only response score. The reported correlation between localization and crop utility is $r=0.77$. These measurements quantify the association, and the gap, between answer presence and the outcome of reading a candidate.

Table~\ref{tab:selector} reports the separate 1,605-question selector analysis. The answer-presence reference selects a candidate containing the reference answer when available. The realized-utility reference selects by the reader's observed candidate score. These reference-guided selections use test-answer information to characterize the evaluated pool. The learned utility scorer produces a different operating point, pairing higher answer scores with wider observations.

\begin{table}[t]
\caption{\textbf{Selector analysis on 1,605 held-out questions.} Reference-guided rows use test-answer information for analysis only. $B$ is mean source-image area.}
\label{tab:selector}
\centering\smalltable
\begin{tabular}{@{}lrr@{}}
\toprule
Selector & ANLS & $B$\\
\midrule
Random candidate & 0.383 & 0.169\\
\qcg\ top-1 & 0.474 & 0.064\\
\qcg\ top-2 & 0.621 & 0.163\\
Learned utility scorer & 0.798 & 0.460\\
\midrule
Answer-presence reference & 0.739 & 0.130\\
Realized-utility reference & 0.903 & 0.370\\
\bottomrule
\end{tabular}
\end{table}

\subsection{Structure-Only Probe}
\label{app:structure}
Table~\ref{tab:structure} compares inputs on 600 DocVQA questions. OCR/layout serialization supplies recognized text and relations. Anchor-only serialization keeps matched lines; type-only and geometry-only variants remove literal text. The comparison shows that lexical content and broader context carry much of the structural input's signal. Pixel-crop and full-image conditions provide visual observations to the same answering model.

\begin{table}[t]
\caption{\textbf{Structure-only probe.} ANLS on 600 DocVQA validation questions. Serialization rows supply OCR/layout text inputs.}
\label{tab:structure}
\centering\smalltable
\begin{tabular}{@{}lr@{}}
\toprule
Input & ANLS\\
\midrule
Full-image perception & 0.890\\
\qcg\ pixel crop & 0.736\\
Full OCR/layout serialization & 0.684\\
Anchor-matched lines only & 0.284\\
OCR type structure & 0.029\\
Layout geometry only & 0.033\\
Shuffled structure & 0.062\\
\bottomrule
\end{tabular}
\end{table}

\subsection{Same Page, Different Questions}
\label{app:samepage}
Across 11,857 pairs of DocVQA questions sharing a page, mean top-1 window intersection-over-union is 0.244, and 65.7\% have overlap below 0.1. Mean normalized center distance is 0.179. Comparing true-question and shuffled-question windows yields mean overlap 0.287 across 5,349 examples. These coordinate measurements complement the answer-score controls in Section~\ref{sec:query_results}.

Figure~\ref{fig:samepage} shows four pages selected by a fixed rule: retain pages with at least three correctly answered questions, order by question count and then document identifier, and display the first four. Each color marks one question's window on the same cached graph.

\subsection{Paired Native Self-Zoom Comparison}
\label{app:native}
Table~\ref{tab:native} summarizes the paired \vstar\ policies. Their accuracy difference is $+0.031$, with a 95\% confidence interval of $[-0.031,+0.094]$ and exact McNemar $p=0.43$. Mean window overlap is 0.53. There are 136 questions answered correctly by both policies, 17 only by self-zoom, 23 only by \qcg, and 15 by neither. The hindsight either-correct score is therefore $176/191=0.921$ after rounding.

\begin{table}[t]
\caption{\textbf{Paired \vstar\ policies.} Accuracy and observation extent with the same crop reader ($n=191$).}
\label{tab:native}
\centering\smalltable
\begin{tabular}{@{}lrr@{}}
\toprule
Quantity & Self-zoom & \qcg\ top-2\\
\midrule
Accuracy & 0.801 & 0.832\\
Mean area & 0.086 & 0.191\\
Invalid coordinates & 2.1\% & 0\%\\
No-detection fallback & --- & 3.7\%\\
Attribute accuracy & 0.878 & 0.817\\
Relation accuracy & 0.684 & 0.855\\
\bottomrule
\end{tabular}
\end{table}

On a separate 500-question InfographicVQA slice, native self-zoom reaches 0.527 ANLS. Syntactically invalid boxes occur in 15.8\% of examples and trigger full-view fallback; valid-box examples attain 0.483. The structural-policy configurations reach 0.651--0.686 on that slice at comparable mean area. These dense-page examples provide a second setting in which to compare coordinate sources.

\subsection{Retrospective Observation Regimes}
\label{app:regimes}
Figure~\ref{fig:regimes} displays performance relative to the same reader's high-resolution full-view score. Best-variant bars summarize a post-hoc envelope of unrefined configurations, and hatching identifies mixed crop/full-view policies. Two control-normalized gaps summarize sensitivity to the selected evidence:
\begin{equation}
 \Delta_q=\frac{s_{\mathrm{top1}}-s_{\mathrm{shuffle}}}{s_{\mathrm{full}}},
 \qquad
 \Delta_{\mathrm{loc}}=\frac{s_{\mathrm{top1}}-s_{\mathrm{anti}}}{s_{\mathrm{full}}}.
 \label{eq:gaps}
\end{equation}
Here $s$ is the benchmark score of the indicated observation policy. The first gap measures sensitivity to the question, and the second to a displaced observation. Table~\ref{tab:regimes} also reports the positive-anchor rate and median source maximum side divided by the 1,536-pixel full-view render cap. TextVQA gaps use its anchored subset.

\begin{table}[t]
\caption{\textbf{Retrospective indicators.} Resolution is median source maximum side divided by 1,536. Anchor is the positive-anchor/detection rate. TextVQA control gaps use anchored questions.}
\label{tab:regimes}
\centering\smalltable
\begin{tabular}{@{}lrrrr@{}}
\toprule
Benchmark & $\Delta_q$ & $\Delta_{\mathrm{loc}}$ & Resolution & Anchor\\
\midrule
ChartQA & 0.01 & 0.13 & 0.52 & 0.77\\
OCRBench & 0.10 & 0.16 & 0.33 & 0.37\\
TextVQA & 0.14 & 0.16 & 0.67 & 0.26\\
DocVQA & 0.31 & 0.48 & 1.46 & 0.82\\
InfographicVQA & 0.13 & 0.16 & 1.59 & 0.98\\
\vstar & 0.25 & 0.57 & 1.46 & 0.96\\
\bottomrule
\end{tabular}
\end{table}

\subsection{No-Anchor Behavior on TextVQA}
\label{app:fallback_audit}
All counts in Table~\ref{tab:no_anchor} use the same 4,850-question evaluation subset. There are 3,613 questions without a positive anchor (74.5\%). The ungated policy usually expands a small structural default, giving an overall top-1 mean area of 0.126. Its no-anchor windows coincide with shuffled-question windows on 3,401 cases (94.1\%). The shared default explains the weak query dependence on this subset.

\begin{table}[t]
\caption{\textbf{No-anchor observations on TextVQA.} Counts use the 4,850-question held-out set. Full-image extent refers to the final window.}
\label{tab:no_anchor}
\centering\smalltable
\begin{tabular}{@{}>{\raggedright\arraybackslash}p{0.79\columnwidth}r@{}}
\toprule
Quantity & Count\\
\midrule
All evaluation questions & 4,850\\
No positive anchor & 3,613\\
No-anchor window equals shuffled window & 3,401\\
No-anchor window has full-image extent & 63\\
Of these, cases with zero OCR lines & 50\\
\bottomrule
\end{tabular}
\end{table}

Among the 63 no-anchor observations with full-image extent, 50 have no OCR lines. The gate takes a different decision: every no-anchor question uses the full view. The question-only ablation also falls back on no match, accounting for its 0.754 mean area and 0.594 accuracy. These branches explain the observation sizes of the corresponding conditions.

\section{Qualitative Evidence and Failure Analysis}
\label{app:qualitative}
\subsection{Selection Rules}
\begin{figure*}[p]
\centering
\includegraphics[width=\textwidth]{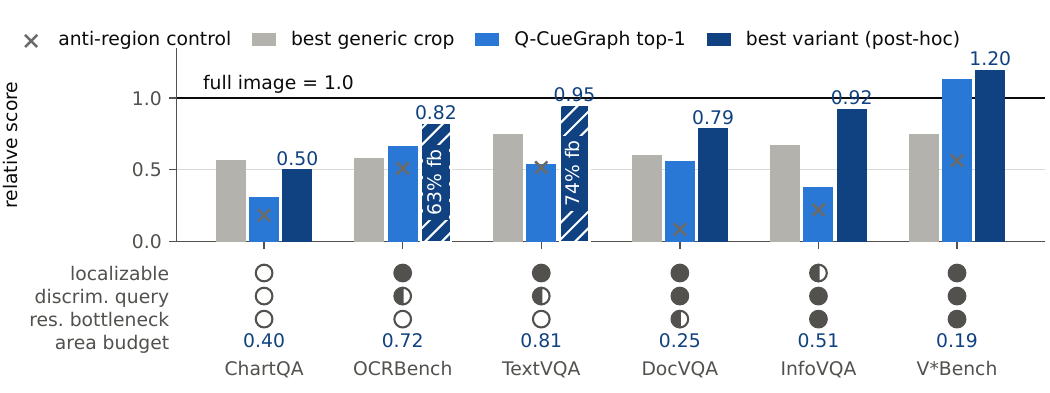}
\caption{\textbf{Cross-benchmark observation regimes.} Scores are normalized by the same reader's high-resolution full-image score. Best-variant bars show the post-hoc unrefined envelope; hatched bars include full-view fallback. Lower markers qualitatively describe evidence localizability, query discrimination, and resolution pressure (open: weak; partial: intermediate; filled: strong). Complete scores and areas appear in Tables~\ref{tab:grid_doc}--\ref{tab:grid_vstar}.}
\label{fig:regimes}
\vspace{14pt}
\includegraphics[width=0.94\textwidth]{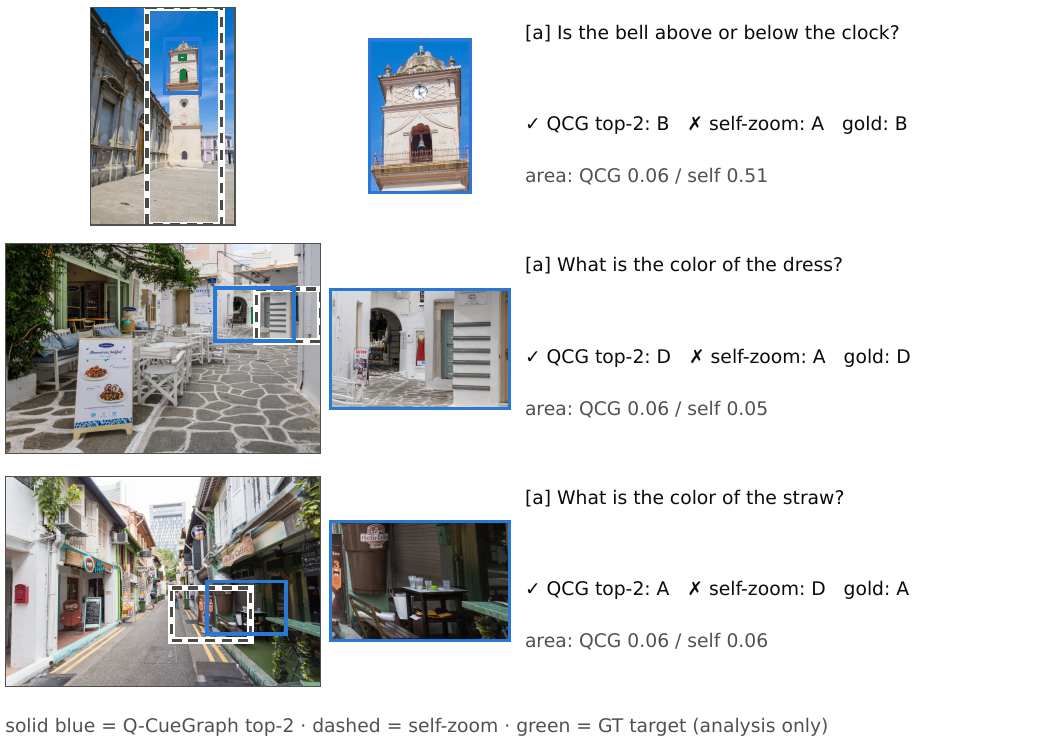}
\caption{\textbf{\qcg\ correct, native self-zoom incorrect.} The cases follow the area-ordered rule in Appendix~\ref{app:qualitative}. Solid blue: \qcg\ top-2; dashed: native self-zoom; green: reference targets shown for analysis.}
\label{fig:pairs_a}
\end{figure*}

\begin{figure*}[p]
\centering
\includegraphics[width=0.94\textwidth]{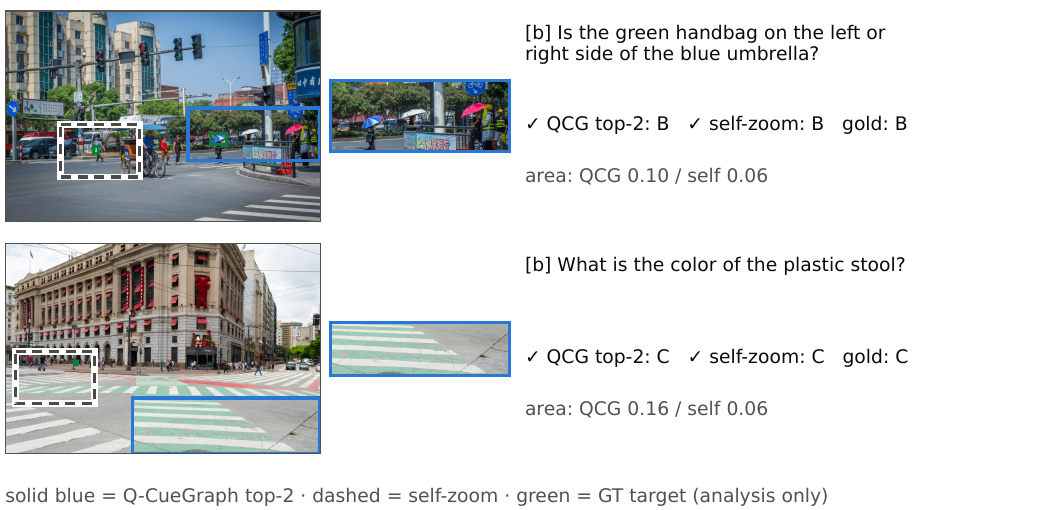}
\caption{\textbf{Both policies correct with different windows.} Cases are ordered by the smallest overlap among both-correct windows. Distinct observations can support the same answer.}
\label{fig:pairs_ab}
\vspace{14pt}
\includegraphics[width=0.94\textwidth]{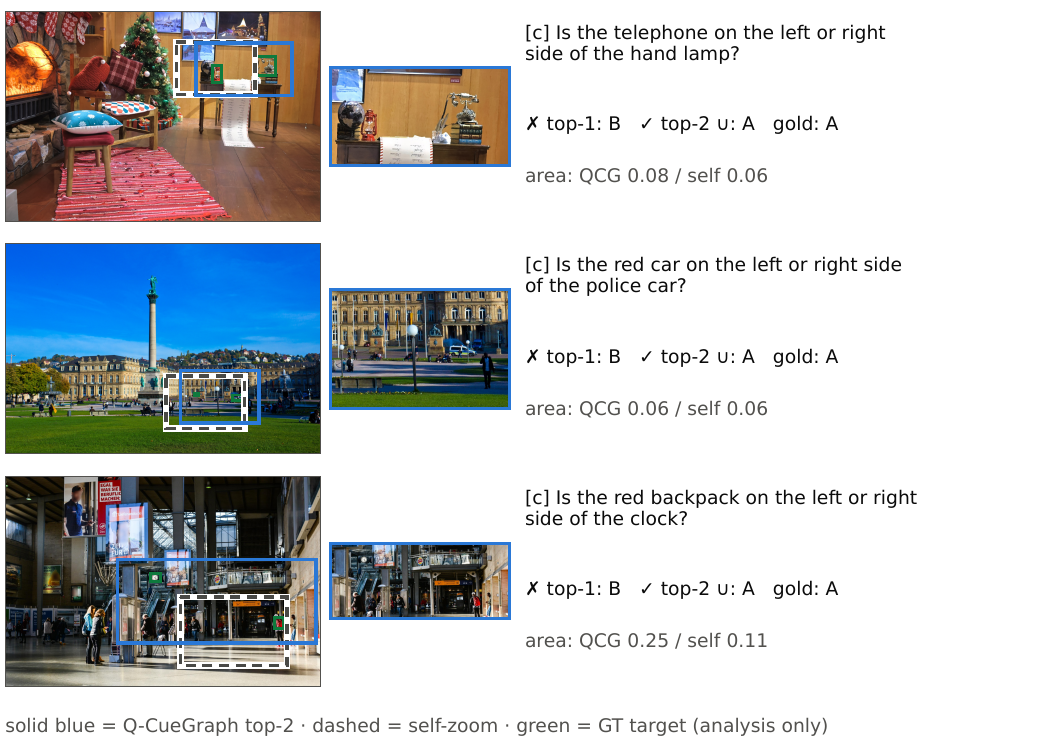}
\caption{\textbf{Composition for relations.} Top-1 is incorrect and the top-2 union supports the correct relation in these recorded cases. Displayed area pairs compare \qcg\ with native self-zoom. Green boxes show reference targets for analysis.}
\label{fig:pairs_b}
\end{figure*}

Paired \vstar\ examples follow fixed selection rules over recorded predictions. Figure~\ref{fig:pairs_a} shows the first three cases, ordered by increasing \qcg\ top-2 area, where self-zoom is incorrect and \qcg\ is correct. Figure~\ref{fig:pairs_ab} shows the first two both-correct cases with the smallest overlap between their windows. Figure~\ref{fig:pairs_b} shows the first three relative-position questions, in identifier order, for which top-1 is incorrect and top-2 is correct. Reference target boxes are post-hoc overlays. Aggregate counts appear in Appendix~\ref{app:native}.

\subsection{Localization and Utility Cases}
Figure~\ref{fig:locutil_cases} illustrates four outcomes: a localized crop supporting a correct scored answer, a localized crop receiving an incorrect score, a ranking miss recovered by refinement, and an error despite target coverage. The second case contains the reference entity in a longer response. It exemplifies the distinction between answer-string presence and the response required by the evaluation protocol. Predictions in the panels are abbreviated for display.

\subsection{Failure Categories}
Manual categorization of anchored TextVQA failures assigns 30\% to missing OCR evidence, 27\% to candidate selection, and 43\% to the scored reader response after selection. Table~\ref{tab:failures} connects these categories to possible interventions. Figure~\ref{fig:failures} shows the recorded examples. The categories locate different points in evidence acquisition: extraction determines available content, selection determines what is presented, and the reader interprets the resulting observation. Globally distributed chart evidence motivates retaining a wider or complete view.

\begin{table}[t]
\caption{\textbf{Failure categories on anchored TextVQA.} Shares are computed over failures of the padded gated policy. ChartQA supplies the additional global-evidence category.}
\label{tab:failures}
\centering\smalltable
\begin{tabular}{@{}>{\raggedright\arraybackslash}p{0.37\columnwidth}r>{\raggedright\arraybackslash}p{0.39\columnwidth}@{}}
\toprule
Category & Share & Candidate intervention\\
\midrule
Missing extracted evidence & 30\% & Improve extraction or fall back\\
Selection failure & 27\% & Rerank or expand candidates\\
Reader-response failure & 43\% & Inspect context, recognition, and scoring\\
Global-evidence mismatch & --- & Retain a wider or full view\\
\bottomrule
\end{tabular}
\end{table}

\fullfigure{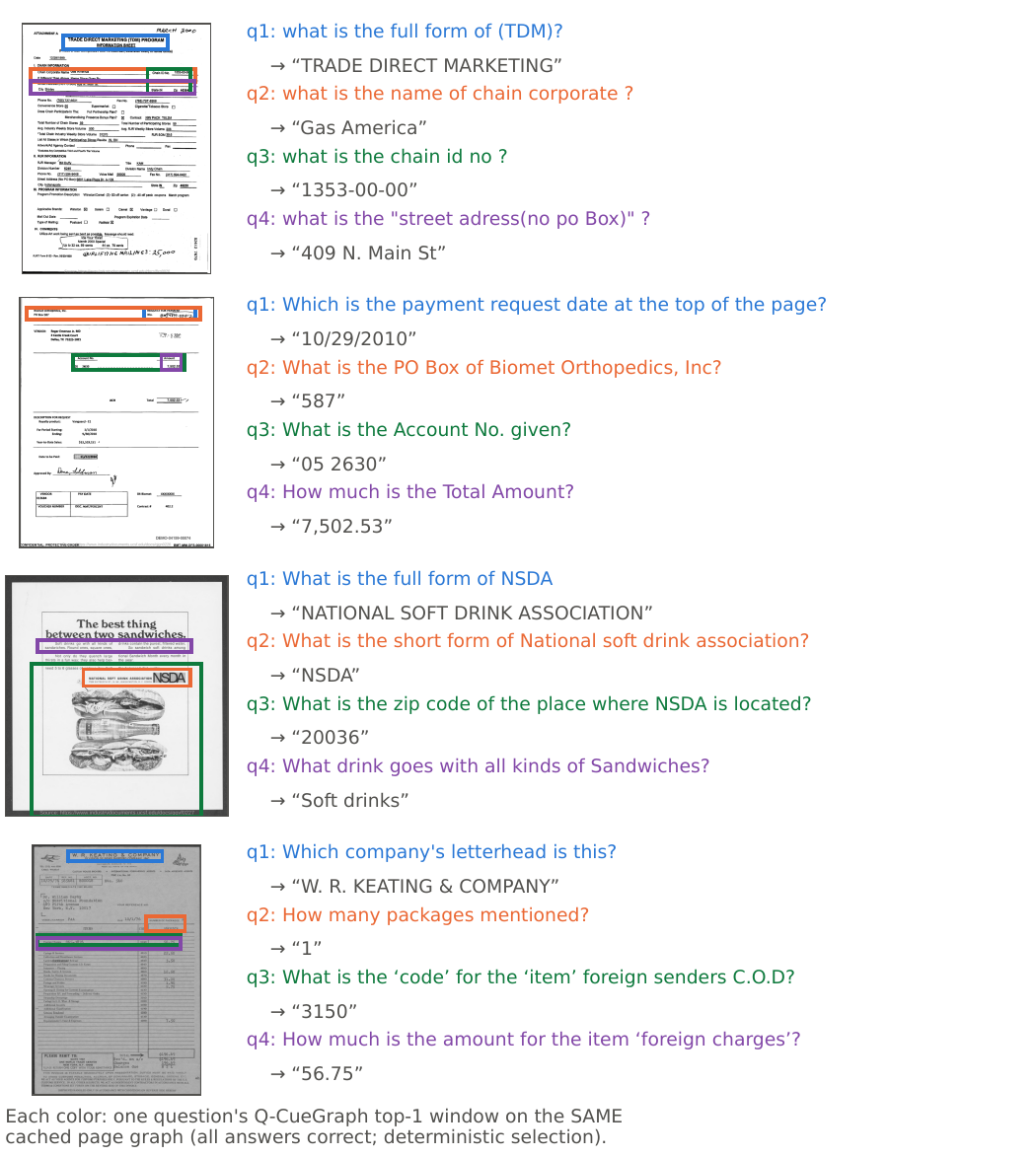}{\textbf{One cached page graph, several observations.} Each color denotes a question's top-1 window. The pages follow the selection rule in Appendix~\ref{app:samepage}. Displayed answers are correct under the recorded evaluation; the graph is reused while the selected evidence changes.}{fig:samepage}
\fullfigure{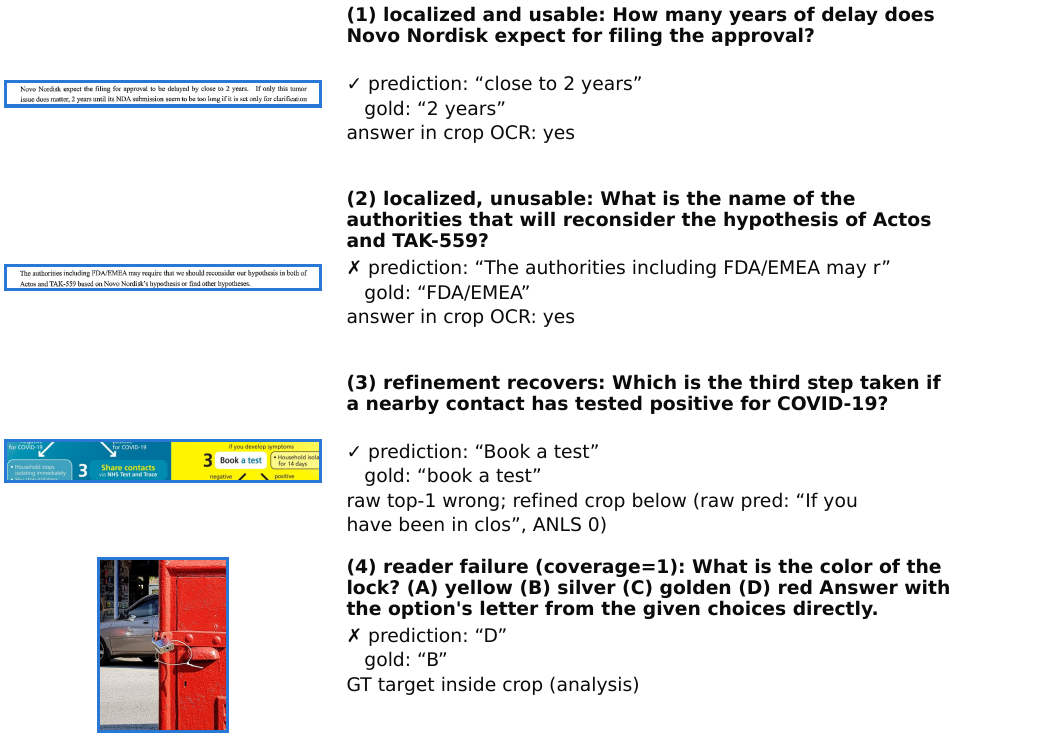}{\textbf{Localization and scored utility.} Crops and responses illustrate answer presence, utility failure despite presence, recovery by refinement, and failure despite target coverage. Displayed predictions are abbreviated.}{fig:locutil_cases}
\fullfigure{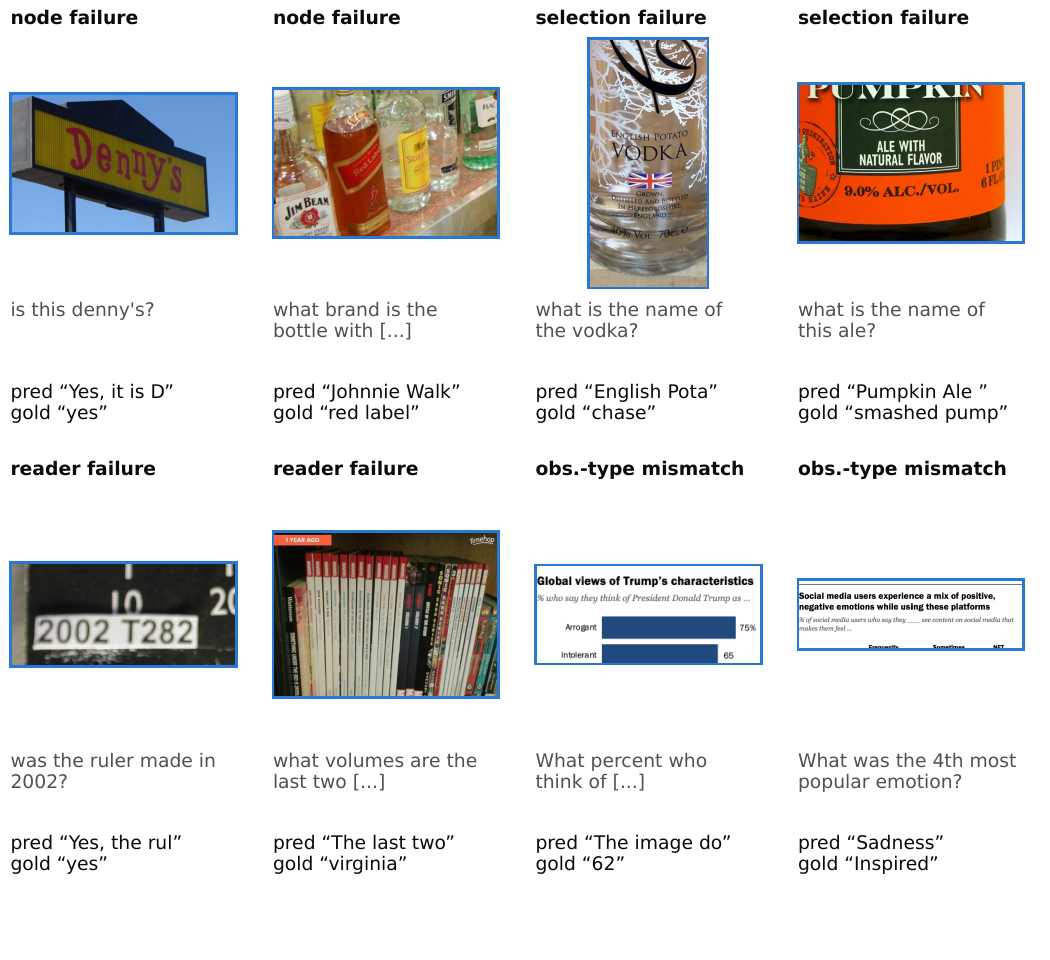}{\textbf{Recorded failure examples.} Diagnostic labels distinguish extraction, selection, scored reader response, and globally distributed evidence. Predictions are abbreviated for display.}{fig:failures}
\FloatBarrier

\end{document}